\documentclass[runningheads]{llncs}

\usepackage[dvipsnames, table]{xcolor}
\usepackage{eccv}

\usepackage{eccvabbrv}

\usepackage{graphicx}
\usepackage{booktabs}

\usepackage[accsupp]{axessibility}  

\usepackage{hyperref}

\usepackage{orcidlink}

\definecolor{green}{rgb}{0.886, 0.965, 0.792}
\definecolor{orange}{rgb}{1.0, 0.902, 0.780}
\definecolor{blue}{rgb}{0.855, 0.890, 0.953}
\definecolor{myblue}{RGB}{0,0,255}
\definecolor{mycolor1}{RGB}{101, 183, 65}
\definecolor{mycolor2}{RGB}{255, 164, 71}
\definecolor{mycolor3}{RGB}{0, 169, 255}

\usepackage{multirow}
\usepackage{tabularx}
\usepackage{array}

\usepackage{amssymb}
\makeatletter
\newcommand{\rmnum}[1]{\romannumeral #1}
\newcommand{\Rmnum}[1]{\expandafter\@slowromancap\romannumeral #1@}
\makeatother

\usepackage{pifont}
\newcommand{\cmark}{\texttt{\ding{51}}}

\newcolumntype{x}[1]{>{\centering\arraybackslash}p{#1pt}}
\newcolumntype{y}[1]{>{\raggedright\arraybackslash}p{#1pt}}
\newcolumntype{z}[1]{>{\raggedleft\arraybackslash}p{#1pt}}

\definecolor{baselinecolor}{gray}{0.8}
\usepackage{subcaption}

\usepackage{wrapfig}
\usepackage{marvosym}

\begin{document}

\title{AUFormer: Vision Transformers are Parameter-Efficient Facial Action Unit Detectors} 

\titlerunning{AUFormer}

\author{Kaishen Yuan\inst{1,2}\thanks{Equal contribution~~~\textsuperscript{\Letter} Corresponding authors: Xin Liu and Zitong Yu } \and
Zitong Yu\inst{2}$^{\star}$\textsuperscript{(\Letter)} \and
Xin Liu\inst{3}\textsuperscript{(\Letter)} \and
Weicheng Xie\inst{4} \and \\
Huanjing Yue\inst{1} \and
Jingyu Yang\inst{1}}

\authorrunning{K.~Yuan et al.}

\institute{Tianjin University \and
Great Bay University \and
Lappeenranta-Lahti University of Technology \and
Shenzhen University \\
\email{\{yuankaishen,huanjing.yue,yjy\}@tju.edu.cn, yuzitong@gbu.edu.cn, \\
linuxsino@gmail.com, wcxie@szu.edu.cn}}

\maketitle

\begin{abstract}
Facial Action Units (AU) is a vital concept in the realm of affective computing, and AU detection has always been a hot research topic. Existing methods suffer from overfitting issues due to the utilization of a large number of learnable parameters on scarce AU-annotated datasets or heavy reliance on substantial additional relevant data. Parameter-Efficient Transfer Learning (PETL) provides a promising paradigm to address these challenges, whereas its existing methods lack design for AU characteristics. Therefore, we innovatively investigate PETL paradigm to AU detection, introducing \textbf{AUFormer} and proposing a novel Mixture-of-Knowledge Expert (MoKE) collaboration mechanism. An individual MoKE specific to a certain AU with minimal learnable parameters first integrates personalized multi-scale and correlation knowledge. Then the MoKE collaborates with other MoKEs in the expert group to obtain aggregated information and inject it into the frozen Vision Transformer (ViT) to achieve parameter-efficient AU detection. Additionally, we design a Margin-truncated Difficulty-aware Weighted Asymmetric Loss (MDWA-Loss), which can encourage the model to focus more on activated AUs, differentiate the difficulty of unactivated AUs, and discard potential mislabeled samples. Extensive experiments from various perspectives, including within-domain, cross-domain, data efficiency, and micro-expression domain, demonstrate AUFormer's state-of-the-art performance and robust generalization abilities without relying on additional relevant data. The code for AUFormer is available at \url{https://github.com/yuankaishen2001/AUFormer}.
\end{abstract}

\section{Introduction}
\label{sec:intro}
\begin{wrapfigure}{r}{0.45\textwidth}
    \centering
    \includegraphics[width=1.0\linewidth]{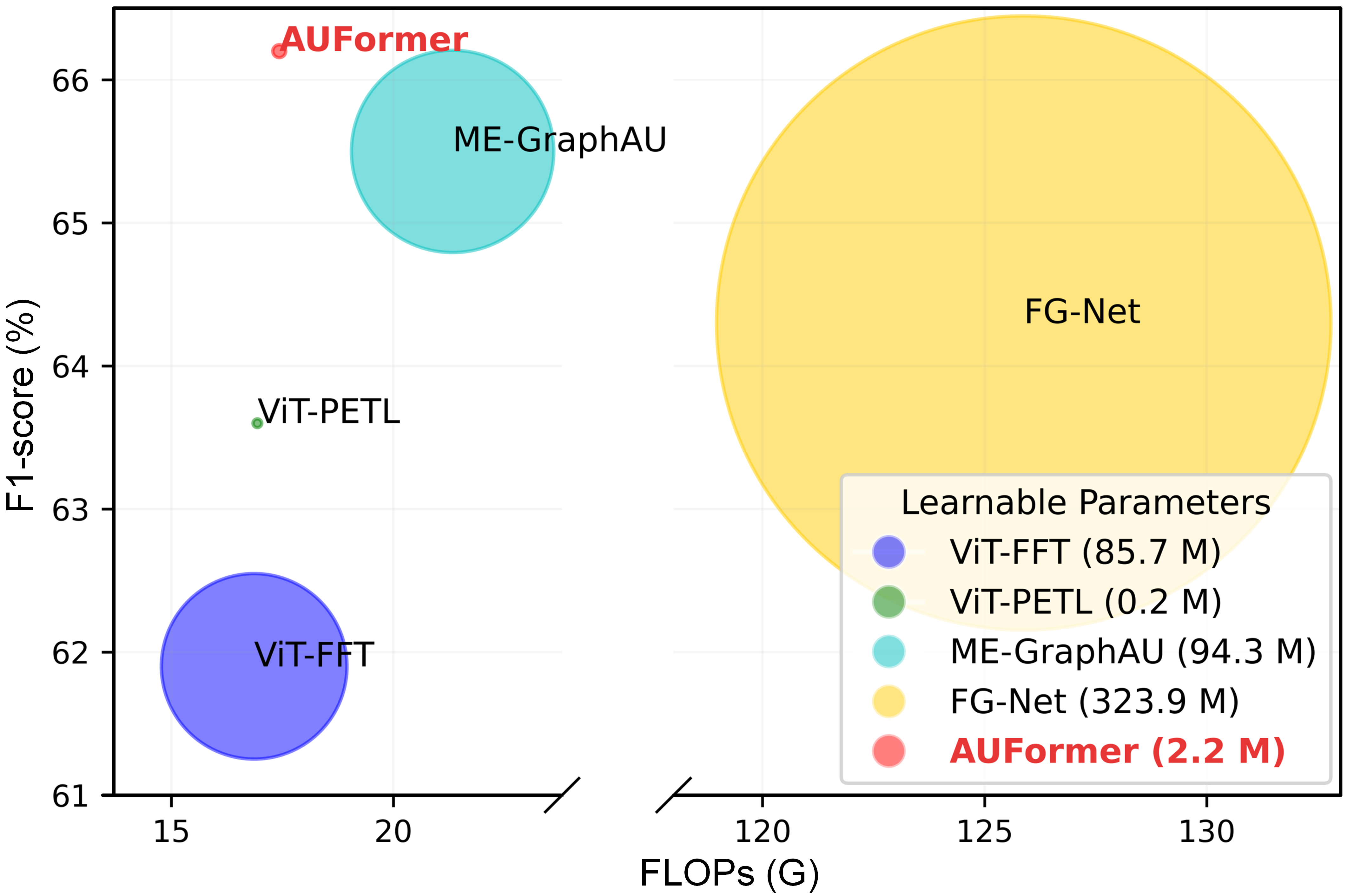}
    \vspace{-2.0em}
    \caption{
    Comparison of AUFormer with fully fine-tuning, PETL paradigm, and the state-of-the-art methods on BP4D~\cite{BP4D} in terms of learnable parameters, FLOPs, and F1-score. The size of bubbles is proportional to the number of learnable parameters. 
    }
    \label{fig.1}
    \vspace{-2.6em}
\end{wrapfigure}

Facial Action Units (AU), defined by the Facial Action Coding System (FACS)~\cite{FACS}, delineate the movements or deformations of facial muscles and can be utilized to analyze intricate facial expressions. Consequently, AU detection has become a fervently researched topic in recent years, and a large number of powerful AU detection methods have surfaced~\cite{AUsurvey}. Considering the varied sizes of facial regions corresponding to different AUs and anatomical potential co-occurrences or mutual exclusivity between them, existing methods enhanced AU detection performance by designing diverse multi-scale features~\cite{DRML,JAA-Net,PIAP,FG-Net,EAC-Net} or modeling correlation information~\cite{SRERL,HMP-PS,LP-Net,UGN-B}, respectively. Most of these methods relied on Convolutional Neural Network (CNN)~\cite{ResNet} or Graph Neural Network (GNN)~\cite{GGNN}, which limit the model's perspective to local regions. Some emerging methods~\cite{MPSCL,ME-GraphAU,FAUDT,FAN-Trans,SLFAUD} introduced Transformer~\cite{Attention}, leveraging its powerful capability to capture long-range dependencies for extracting richer global context information from facial data. Unfortunately, the methods mentioned above inevitably encounter overfitting issue arising from fully fine-tuning pre-trained models with a large number of learnable parameters on scarce AU-annotated data. Recently, some methods have explored using self-supervised learning~\cite{MAE-AU,KDSRL} or incorporating auxiliary information~\cite{BG-AU,SEV-Net,MCM} to mitigate the issue of overfitting; however, substantial amounts of additional relevant data are still necessary~\cite{CelebA,AffectNet,BP4D+,FACS,3DMesh}. Therefore, \textit{how to efficiently learn reliable AU representations on limited AU-annotated data remains a crucial problem to be addressed.}

Parameter-Efficient Transfer Learning (PETL) presents a promising strategy to alleviate the overfitting or catastrophic forgetting issues that arise when fully fine-tuning a pre-trained model due to insufficient data for downstream tasks. Recently, PETL paradigm has demonstrated strong capabilities in efficiently fine-tuning models pre-trained on large-scale datasets~\cite{Adapter,LoRA,VPT,Convpass,NOAH,AdaptFormer}. Concretely, PETL methods insert lightweight modules into the pre-trained model and, during the training process, freezing the pre-trained parameters while only fine-tuning these added modules end-to-end, adapting the model to the downstream task. Although PETL methods could be beneficial for AU detection, direct utilization of existing adaptation modules is suboptimal. This is because current PETL methods do not consider the characteristics of AUs, namely, multi-scale features and correlation information. Naturally, \textit{designing an adaptation module specialized for AU detection is a research direction worth exploring.}

Based on the observations above, we are the first to investigate the PETL paradigm to AU detection, introducing \textbf{AUFormer}, and innovatively propose a Mixture-of-Knowledge Expert (MoKE) collaboration mechanism to efficiently leverage pre-trained Vision Transformer (ViT~\cite{ViT}). Specifically, we freeze the pre-trained ViT and introduce a set of MoKEs with minimal learnable parameters for each AU to efficiently mine personalized features specific to a certain AU. The MoKEs integrate Multi-Receptive Field (MRF) and Context-Aware (CA) operators to obtain multi-scale and correlation knowledge that play a key role in AU detection. At each layer, MoKEs for individual AUs, enriched with knowledge, collaborate within their expert group to aggregate information and inject it back into ViT, enabling ViT to adapt to AU detection. As shown in Figure \ref{fig.1}, \textit{AUFormer is parameter-efficient, achieving superior performance by fine-tuning only a small number of parameters without using any additional relevant data.} \textbf{(\rmnum{1})} Compared to fully fine-tuning ViT (ViT-FFT)~\cite{ViT}, AUFormer achieves significantly better performance with fewer learnable parameters and a slight increase of 3\% in FLOPs, owing to PETL paradigm. \textbf{(\rmnum{2})} Compared to existing PETL method (ViT-PETL, taking Convpass~\cite{Convpass} as an example), benefiting from the tailored MoKEs and their efficient intra-group collaboration, AUFormer achieves substantial improvement with a manageable computational overhead. \textbf{(\rmnum{3})} Compared to the state-of-the-art methods, ME-GraphAU~\cite{ME-GraphAU} and FG-Net~\cite{FG-Net}, AUFormer outperforms them with only 2.3\% and 0.7\% of learnable parameters and fewer FLOPs. 

Additionally, the supervision signal is also a key factor affecting model performance, and we notice that commonly utilized loss functions, such as weighted multi-label cross-entropy loss (WCE-Loss)~\cite{DRML,JAA-Net,FAUDT,LP-Net} and weighted asymmetric loss (WA-Loss)~\cite{ALoss,ME-GraphAU}, do not fully consider the nature of AU datasets and the differences in difficulty among AUs. \textit{The development of a superior supervision signal is imminent.} Therefore, we design a novel Margin-truncated Difficulty-aware Weighted Asymmetric Loss function (MDWA-Loss), which focuses more on activated AUs, differentiates the difficulty of unactivated AUs, and discards potentially mislabeled samples, encouraging model to learn more valuable information. 

We conduct extensive experiments on macro-expression AU datasets, namely BP4D~\cite{BP4D} and DISFA~\cite{DISFA}, showcasing AUFormer's robustness and generalizability superiority over state-of-the-art methods from multiple perspectives, including within-domain, cross-domain, and data efficiency. Furthermore, we evaluate AUFormer on the micro-expression AU dataset (CASME \Rmnum{2}~\cite{CASME}), demonstrating its effectiveness even in the micro-level domain.

Our contributions can be summarized as follows:
\begin{itemize}
    \item We are the first to investigate the PETL paradigm to AU detection, proposing the AUFormer and developing a parameter-efficient MoKE collaboration mechanism.
    \item We design tailored MoKEs, which integrate MRF and CA operators, for respectively extracting multi-scale and correlation knowledge essential for detecting AU.
    \item We introduce a novel MDWA-Loss that focuses more on activated AUs, differentiates the difficulty of unactivated AUs, and discards potentially mislabeled samples.
    \item We conduct extensive experiments on BP4D and DISFA, showcasing AUFormer's superior robustness and generalization from multiple perspectives. Besides, we validate its effectiveness in a micro-expression domain (CASME \Rmnum{2}), a novel evaluation not attempted by previous methods.
\end{itemize}

\section{Related Work}
\label{sec:related work}

\textbf{Facial Action Unit Detection.}
With the rapid development of Computer Vision (CV) technology in recent years, more and more researchers are devoted to AU detection. Most traditional methods relied on hand-crafted features, leading to limited performance~\cite{ginosar2015century,jiang2011action,zhao2015joint,wang2013capturing}. The rise of deep learning has greatly promoted the progress of AU detection. Due to the sizes of the muscles associated with different AUs are not consistent, some methods focused on learning multi-scale features to enhance the model’s ability~\cite{DRML,JAA-Net,PIAP,FG-Net,EAC-Net}. J{\^A}A-Net~\cite{JAA-Net} introduced the hierarchical and multi-scale region layer, which divided feature map into several patches and applied independent convolutional kernels to each patch. The patch sizes increased layer by layer to extract multi-scale information. From another perspective, considering intricate inherent correlations among AUs, some methods were dedicated to studying effective modeling of correlation information~\cite{ME-GraphAU,MPSCL,FAN-Trans,FAUDT,LP-Net,HMP-PS,SRERL}. ME-GraphAU~\cite{ME-GraphAU} proposed a graph structure with multi-dimensional edge features, and used a Gated Graph Convolutional Network (GGCN)~\cite{GGCN} to capture more abundant intrinsic correlation clues for each AU pair. Most existing methods were based on CNN or GNN, limiting model’s perspective to local regions, while some emerging methods~\cite{MPSCL,FAN-Trans,FAUDT} introduced Transformer to leverage its powerful long-range dependency modeling capabilities. FAUDT~\cite{FAUDT} used Transformer blocks to adaptively extract global correlation knowledge between AUs. Although the methods mentioned above have shown satisfactory results, existing models with a large number of learnable parameters inevitably face the issue of overfitting on limited AU-annotated data. Recently, KDSRL~\cite{KDSRL} and BG-AU~\cite{BG-AU} have attempted to mitigate overfitting through contrastive learning or biomechanical guidance, respectively, but they still necessitates a large amount of additional relevant data. 

Unlike previous methods that utilized Transformer as a backbone~\cite{ME-GraphAU,MPSCL} or for global correlation learning~\cite{FAUDT,FAN-Trans} in a fully fine-tuning paradigm, we propose a novel parameter-efficient MoKE collaboration mechanism to leverage pre-trained ViT. In this mechanism, several experts, each with a small number of learnable parameters and capable of learning personalized features for different AUs, collaborate within the group and inject consolidated knowledge into frozen ViT to adapt it to AU detection. Even though only trained on a limited dataset with AU annotations, AUFormer achieves state-of-the-art performance.

\textbf{Parameter-Efficient Transfer Learning.}
With the rapid growth in model scales, the PETL paradigm has attracted increasing research interest. Initially applied in Natural Language Processing (NLP) for efficiently fine-tuning models pre-trained on large datasets~\cite{BERT,Attention,CLIP}, it has recently shown remarkable success in CV. Adapter~\cite{AdaptFormer,Adapter} was a bottleneck consisting of a downsampling layer, an upsampling layer, and an activation function (such as GELU~\cite{GELU}) inserted between them, which is connected in series or in parallel in the Multi-head Self-attention (MHSA) or Multi-layer Perceptron (MLP) layer of ViT. LoRA~\cite{LoRA} learned the low-rank approximation of increments to indirectly optimize the Queries and Keys in ViT. VPT~\cite{VPT} prepended a set of learnable prompts to the input, which can be viewed as inserting some learnable pixels into the input space, to adapt the pre-trained ViT to new tasks. Recently, Convpass~\cite{Convpass} introduced convolutional layers into the adaptation module to capture spatial information of images and learn visual inductive bias, demonstrating superior performance. However, existing PETL methods that rely solely on naive learnable prompts, linear or convolutional layers cannot fully extract the meticulous knowledge required for AU detection, so directly applying these methods leads to suboptimal results. To our best knowledge, we are the first to investigate the PETL paradigm to AU detection.

\begin{figure*}[t]
    \centering
    \includegraphics[width=1.0\linewidth]{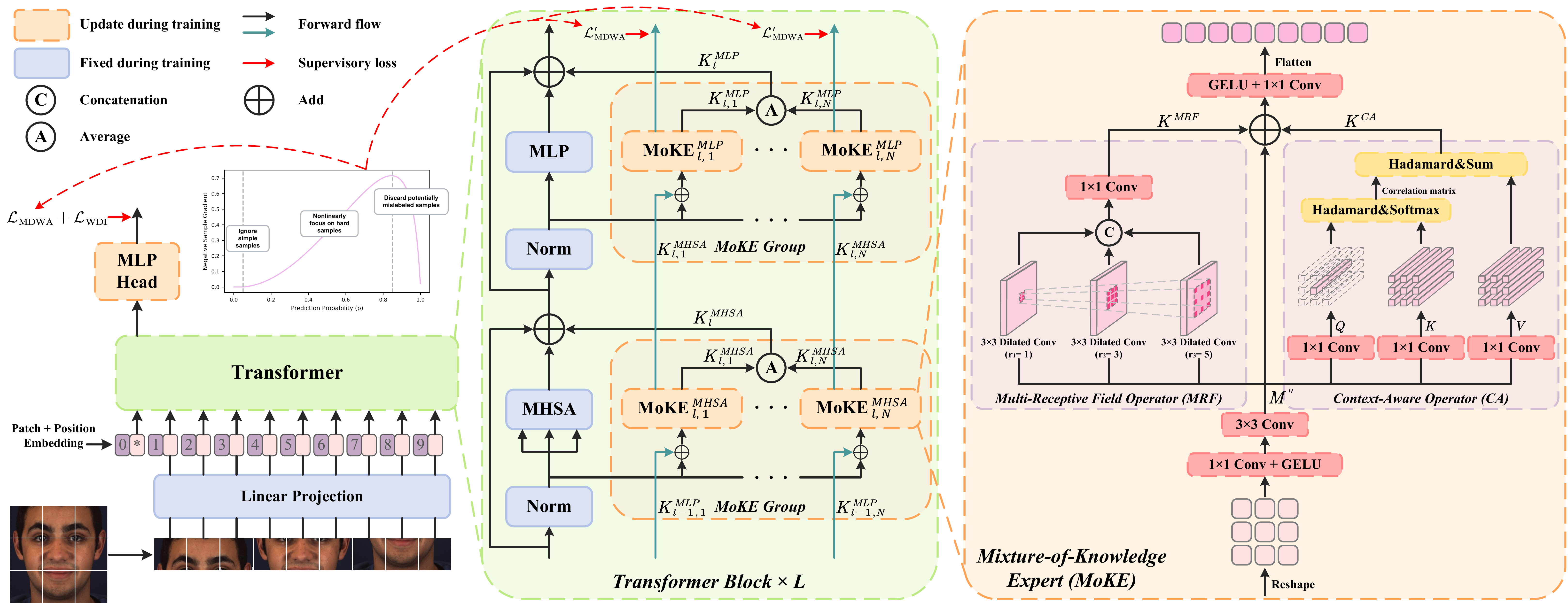}
    \vspace{-2.0em}
    \caption{
    The overall architecture of the proposed \textbf{AUFormer}. MoKEs first extract essential multi-scale and correlation knowledge through MRF and CA operators. Then, the personalized features learned for each AU are integrated through an parameter-efficient intra-group collaboration mechanism for AU detection. We provide a detailed illustration of the $l$-th Transformer block.
    }
    \label{fig.2}
    \vspace{-1.8em}
\end{figure*}

\section{Methodology}
\label{sec:methodology}
\subsection{Preliminary}
We first provide a brief review of ViT~\cite{ViT}. ViT consists of a series of consecutive Transformer blocks, each of which mainly includes MHSA and MLP layers for feature extraction, as shown by the blue squares in Figure \ref{fig.2}. Ignoring the LayerNorm~\cite{LayerNorm} layers, a Transformer block can be represented as 
\begin{equation}
\label{eq:ViT}
    X_{l}^{\prime} = X_{l-1} + \text{MHSA}(X_{l-1}), X_l = X_{l}^{\prime} + \text{MLP}(X_{l}^{\prime}),
\end{equation}
where $X_{l-1}$ and $X_{l}^{\prime}$ respectively represent the tokens input to MHSA and MLP of the $l$-th Transformer block, and $X_{l}$ represents the tokens output from the $l$-th Transformer block. ViT has demonstrated strong feature representation and long-range dependency modeling capabilities. However, fully fine-tuning pre-trained ViT on downstream tasks with limited datasets often leads to overfitting or catastrophic forgetting issues, resulting in suboptimal performance~\cite{Convpass}. Consequently, we proceed to introduce the proposed AUFormer, a Mixture-of-Knowledge Expert (MoKE) collaboration mechanism that can efficiently utilize the pre-trained parameters of ViT.

\subsection{Collaboration Mechanism}
To adapt the pre-trained ViT for AU detection, we freeze the parameters of ViT and introduce $N$ MoKEs with minimal learnable parameters into both MHSA and MLP layers of each Transformer block, forming MoKE groups, where $N$ represents the number of AUs. Each MoKE learns personalized knowledge for a specific AU and receives information from two aspects, namely the global information that will be input to MHSA/MLP and the knowledge inherited from the previous generation of MoKEs. Taking the MoKE group corresponding to MHSA of the $l$-th Transformer block as an example, each MoKE in the group first learns knowledge for the corresponding AU, and this process can be expressed as
\begin{equation}
    K^{MHSA}_{l,i} = \text{MoKE}^{MHSA}_{l,i}(X_{l-1}+K^{MLP}_{l-1,i}),
\end{equation}
where $\text{MoKE}^{MHSA}_{l,i}$ is the MoKE of the $i$-th AU in the MHSA group, $X_{l-1}$ is the input to MHSA, $K^{MLP}_{l-1,i}$ is the output of MoKE of the $i$-th AU in the MLP group of the previous block, $K^{MHSA}_{l,i}$ is the personalized knowledge that $\text{MoKE}^{MHSA}_{l,i}$ has learned for the $i$-th AU.

Then, MoKEs corresponding to different AUs pass on the acquired knowledge to the next generation of MoKEs and collaborate within their group, integrating information by averaging to obtain comprehensive expertise. Specifically, the collaboration process can be represented as
\begin{equation}
    K^{MHSA}_{l} = \frac{1}{N} \sum_{i=1}^{N} K^{MHSA}_{l,i}, 
\end{equation}
where $K^{MHSA}_{l}$ represents comprehensive expertise obtained through collaboration among MoKEs within group.

The comprehensive expertise obtained is injected back into the Transformer block. Therefore, the input of the MLP in the $l$-th block is transformed from \Cref{eq:ViT} to
\begin{equation}
    X_{l}^{\prime} = X_{l-1} + \text{MHSA}(X_{l-1}) + K^{MHSA}_{l}. 
\end{equation}

The green area in Figure \ref{fig.2} illustrates the overall process described above. The process for MoKE groups of MLP is similar and will not be reiterated here.

\subsection{Structure of MoKE}
To ensure that MoKEs can acquire knowledge beneficial for AU detection, we incorporate Multi-Receptive Field (MRF) and Context-Aware (CA) operators to capture multi-scale and correlation knowledge relevant to AUs, as shown in the orange area of Figure \ref{fig.2}. 

Specifically, since ViT flattens the image into 1D token sequences, the input $M \in \mathbb{R}^{N_t \times D}$ of MoKEs is firstly reshaped to get $M^{\prime} \in \mathbb{R}^{H \times W \times D}$ to restore the 2D spatial structure, where $N_t$, $D$, $H$, and $W$ represent the number of tokens, the number of channels, and the height and width of reshaped features, respectively. Then, $M^{\prime}$ undergoes channel reduction through a $1 \times 1$ convolutional layer and is further processed by a $3 \times 3$ convolutional layer to extract basic features $M^{\prime\prime} \in \mathbb{R}^{H \times W \times d}$ which are input into the MRF and CA operator, where $d$ is the number of channels after reduction. It should be noted that we treat the [CLS] token as an individual image following \cite{Convpass}. Due to the identical process, we omit the description of its transformation here.

\textbf{MRF operator} is designed to extract features with varying receptive field sizes, accommodating the distinct muscle region sizes corresponding to different AUs. It consists of three parallel dilated convolutional layers, each with a $3 \times 3$ kernel, where the dilated rates increase gradually, denoted as $r_1$, $r_2$ and $r_3$. The outputs of these three dilated convolutional layers are concatenated and channel-compressed through a $1 \times 1$ convolutional layer to obtain integrated multi-scale knowledge $K^{MRF}$. 

\textbf{CA operator} focuses on perceiving the contextual information of features, thereby implicitly modeling potential correlation knowledge among muscles. Specifically, input basic features $M^{\prime\prime}$ are firstly mapped to $Q$, $K$, and $V$ through three parallel $1 \times 1$ convolutional layers. Then, $Q$ is Hadamard producted with the vectors of $K$ in its neighborhood (assuming a size of $S \times S$) and passed through Softmax to obtain correlation matrix. Utilizing obtained correlation knowledge, $V$ is aggregated within the neighborhood of $Q$ to update it to $K^{CA}$. This process can be expressed as 
\begin{equation}
    K^{CA}_{x} = \sum_{x^{\prime} \in \mathcal{R} (x)} \text{Softmax} ( \frac{Q_{x} \odot K_{x^{\prime}}}{\sqrt{d}} ) \odot V_{x^{\prime}},
\end{equation}
where $\odot$ is Hadamard product, $x$ is the spatial index of query vector $Q_{x}$, and $\mathcal{R} (x)$ is the set of indices for the neighborhood of $Q_{x}$ to be aggregated. The purple area in Figure \ref{fig.2} illustrates the details of MRF and CA operator.

Finally, we integrate the basic feature $M^{\prime\prime}$, multi-scale knowledge $K^{MRF}$, and correlation knowledge $K^{CA}$ together, and expand channels through a $1 \times 1$ convolutional layer to obtain informative mixture-of-knowledge features.

\subsection{Loss Function}
\label{sec:LossFunction}
AU detection is typically regarded as a multi-label binary classification problem. To make final predictions, we transform the [CLS] token from the last block of ViT into $Z \in \mathbb{R}^{N}$ through a linear projector and convert it into probabilities $p$ using the Sigmoid activation function. 

Historically, most methods have used a weighted multi-label cross-entropy loss (WCE-Loss) as supervision signal during training. However, this straightforward setup does not account for the issue of significantly fewer activated AUs compared to unactivated AUs. Therefore, ME-GraphAU~\cite{ME-GraphAU} introduced a weighted asymmetric loss (WA-Loss) to guide model's attention towards both activated and challenging unactivated AUs. Nevertheless, WA-Loss does not differentiate the difficulty levels among different AUs, nor does it consider samples potentially mislabeled. Recognizing these issues, we propose a novel Margin-truncated Difficulty-aware Weighted Asymmetric Loss (MDWA-Loss), which can focus more on activated AUs, adaptively discern the difficulty of unactivated AUs, and discard the potentially mislabeled samples. It can be formulated as
\begin{equation}
    \label{eq:MDWA}
    \mathcal{L}_{\text{MDWA}}=-\frac{1}{N} \sum_{i=1}^{N} \omega_i [y_i \log (p_i) + (1 - y_i) p_{m, i} ^{\gamma_i} \log (1 - p_{m, i})],
\end{equation}
where $y_i$ and $p_i$ represent the ground truth and predictions for the $i$-th AU, respectively. $\omega_i$ is the weight for addressing class imbalance, and it can be formulated as $\omega_i = N (1/r_i) / {\sum_{j=1}^{N} (1/r_j)}$, where $r_i$ is the occurrence rate of the $i$-th AU. $p_{m,i}$ is used to discard potentially mislabeled samples and can be formulated as $p_{m,i} = \max (p_i - m, 0)$, where $m \in [0, 1]$ is the truncation margin. $\gamma_i$ is used to distinguish the different difficulty levels of unactivated AUs and can be formulated as $\gamma_i = B_L + (B_R-B_L) \times r_i$, where $B_L$ and $B_R$ are left and right boundaries of $\gamma_i$ respectively.

\begin{figure}[t]
  \begin{minipage}{0.48\linewidth}
    \centering
    \includegraphics[width=\linewidth]{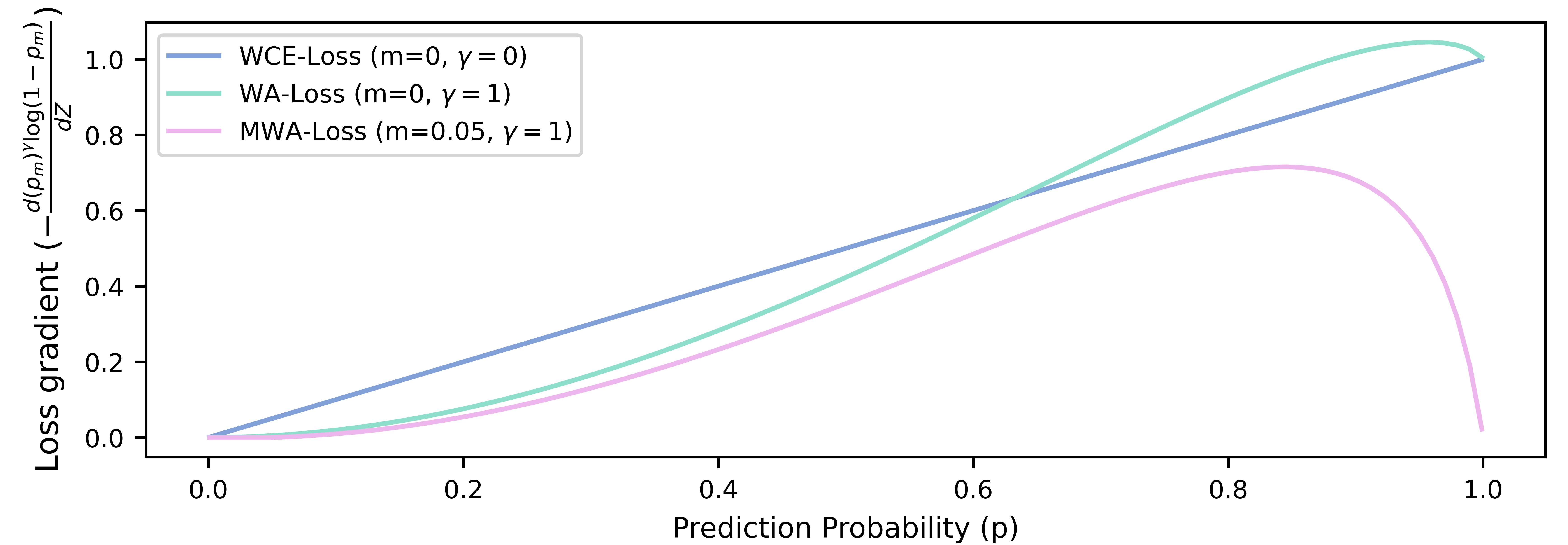}
    \subcaption{Comparison between WCE-Loss, WA-Loss, and MWA-Loss}
    \label{fig:m}
  \end{minipage}
  \hspace{0.1em}
\begin{minipage}{0.48\linewidth}
    \centering
    \includegraphics[width=\linewidth]{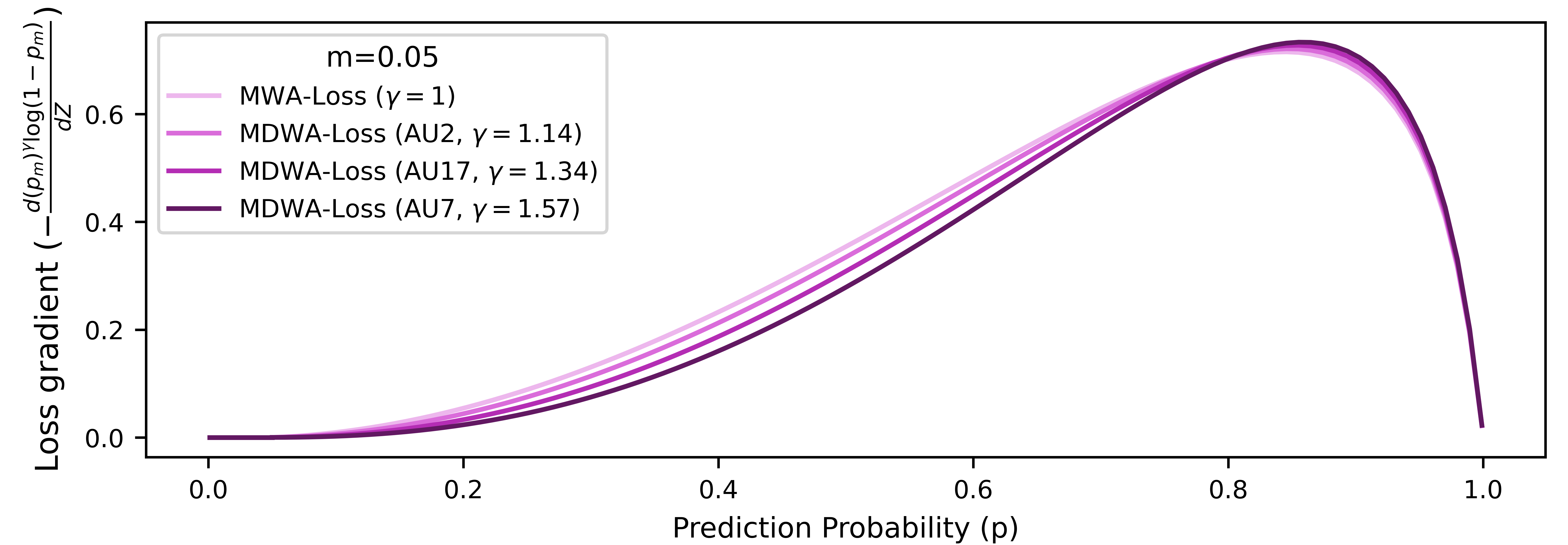}
    \subcaption{Comparison between AUs of different difficulty degrees}
    \label{fig:gamma}
  \end{minipage}
  \vspace{-1.0em}
  \caption{Gradient analysis of unactivated AU (negative sample).}
\label{MDWA-Loss}
\vspace{-1.7em}
\end{figure}

To better understand how MDWA-Loss works, we analyze the gradient of the loss. Ignoring the index $i$ of AUs (e.g., $p_{m, i}$$\rightarrow$$p_{m}$), the loss gradient for unactivated AU, calculated for the output $Z$, can be expressed as
\begin{equation} 
\label{eq:Gradient}
    -\frac{d p_{m} ^{\gamma} \log (1 - p_{m})}{d Z} = 
    \begin{cases} 
    0, &  p < m \\
    p_m ^{\gamma} ( \frac{1}{1-p_m}-\frac{\gamma \log (1-p_m)}{p_m}) p(1-p), &  p \geq m 
    \end{cases}.
\end{equation}
The derivation process is detailed in Appendix. 

We first explore the effect of truncation margin $m$. By fixing $\gamma$ to 1, i.e., temporarily ignoring the difficulty differences of AUs, we can obtain a degenerate version called MWA-Loss. We compare it with the losses commonly used before, and Figure \ref{fig:m} illustrates the gradient curves of different losses. The gradient of WCE-Loss increases linearly and does not focus on the more challenging samples. Although WA-Loss suppresses the gradient of easy negative samples ($p \ll 0.5$), placing more emphasis on the harder samples, its gradient increases with the prediction probability, potentially propagating the gradient of mislabeled samples. In contrast, the MWA-Loss can be divided into three stages. When negative samples are very easy ($p < m$), the gradient is directly ignored. As the difficulty of negative samples increases, the gradient also increases nonlinearly, giving higher attention to challenging samples. When negative samples are extremely difficult ($p \approx 1$), it is suspected that these are actually positive samples that have been mislabeled, so the gradient propagation for such samples is rejected. For densely annotated AU datasets, mislabeled samples are inevitable (several examples are detailed in Appendix), thus the introduction of $m$ is helpful for training.

Then, the role of $\gamma$ is discussed. Since the difficulty of detecting different AUs varies greatly, we introduce $\gamma$ specific to each AU for differentiation. We intuitively use the occurrence rate to measure the difficulty of AUs, and they often show an inverse relationship. Specifically, the lower the occurrence rate, the greater the difficulty, and the smaller the value of $\gamma$, leading to larger gradients being back-propagated. Figure \ref{fig:gamma} shows the $\gamma$ values and gradient curves of AU2, AU7, and AU17, which have different degrees of difficulty, demonstrating the ability of MDWA-Loss to differentiates unactivated AUs.

Besides, referring to \cite{JAA-Net}, we also introduce a weighted multi-label dice loss (WDI-Loss) to alleviate the issue of AU prediction bias towards non-occurrence, as follow
\begin{equation}
    \mathcal{L}_{\text{WDI}} = \frac{1}{N} {\sum_{i=1}^{N} \omega_i (1 - \frac{2 y_i p_i + \varepsilon}{y_i^2 + p_i^2 + \varepsilon})},
\end{equation}
where $\varepsilon$ is a smooth term and the meanings of $y_i$, $p_i$, and $\omega_i$ remain consistent with those defined in \Cref{eq:MDWA}.

Additionally, to ensure that each MoKE can learn personalized knowledge about a particular AU, we pass the [CLS] tokens from the final group of MoKE outputs through linear projectors and apply Sigmoid activation to predict each AU separately. We also employ MDWA-Loss as supervision signal, and denote it as $\mathcal{L}_{\text{MDWA}}^{\prime}$. Due to its expression being the same as $\mathcal{L}_{\text{MDWA}}$, it is omitted here.

The overall loss of AUFormer can be expressed as
\begin{equation}
    \mathcal{L} = \mathcal{L}_{\text{MDWA}} + \mathcal{L}_{\text{WDI}} + 
    \mathcal{L}_{\text{MDWA}}^{\prime}.
\end{equation}

\section{Experimental results}
\label{sec:Experimental results}
\subsection{Settings}
\textbf{Datasets.} We evaluate the proposed AUFormer on three datasets: BP4D~\cite{BP4D} and DISFA~\cite{DISFA} in the macro-expression domain, and CASME \Rmnum{2}~\cite{CASME} in the micro-expression domain. For \textbf{BP4D} and \textbf{DISFA}, following \cite{DRML,ME-GraphAU,JAA-Net}, we employ subject-exclusive 3-fold cross-validation. For \textbf{CASME \Rmnum{2}}, following \cite{ASP}, we extract the apex frame of each video for experiments and employ subject-exclusive 4-fold cross-validation. More details about the datasets and data pre-processing are in Appendix.

\textbf{Implementation Details.}
We choose ViT-B/16 pre-trained on ImageNet-1K~\cite{ImageNet} as the backbone. Regarding the model configuration, the number of channels $d$ after reduction in MoKE is set to 4. The dilated rates $r_1$, $r_2$, and $r_3$ in MRF are set to 1, 3, and 5. The neighborhood size $S \times S$ in CA is set to $3 \times 3$. The left and right boundaries $B_L$ and $B_R$ of $\gamma$ are set to 1 and 2. The truncation margin $m$ for BP4D, DISFA, and CASME \Rmnum{2} are set to 0.1, 0.15, and 0.3. The smooth term $\varepsilon$ is set to 1. The number of AUs $N$ for BP4D, DISFA, and CASME \Rmnum{2} are set to 12, 8 and 8, following \cite{ASP,DRML}. More details about the training are in Appendix.

\textbf{Evaluation Metric.} Consistent with the previous methods~\cite{DRML,EAC-Net,JAA-Net,KS}, F1-score, which is the harmonic mean of precision and recall, is used as a metric to measure AU detection performance.

\subsection{Comparison with State-of-the-Art Methods}

\begin{wraptable}{r}{0.48\textwidth}
\vspace{-3.4em}
\centering
\scriptsize
\caption{Average F1-score (in \%) results for within-domain evaluations on BP4D and DISFA. The best and second best results for each column are bolded and underlined, respectively.}

\label{tab:BP4D&DISFA}
\begin{tabular}{l|c|cc}
\toprule[1pt]
Method & Venue & BP4D & DISFA \\
\midrule
DRML~\cite{DRML} & CVPR 16 & 48.3 & 26.7 \\

LP-Net~\cite{LP-Net} & CVPR 19 & 61.0 & 56.9 \\

SRERL~\cite{SRERL} & AAAI 19 & 62.9 & 55.9 \\

J{\^A}A-Net~\cite{JAA-Net} & IJCV 21 & 62.4 & 63.5 \\

HMP-PS~\cite{HMP-PS} & CVPR 21 & 63.4 & 61.0 \\

SEV-Net~\cite{SEV-Net} & CVPR 21 & 63.9 & 58.8 \\

FAUDT~\cite{FAUDT} & CVPR 21 & 64.2 & 61.5 \\

PIAP~\cite{PIAP} & ICCV 21 & 64.1 & 63.8 \\

KDSRL~\cite{KDSRL} & CVPR 22 & 64.5 & 64.5 \\

ME-GraphAU~\cite{ME-GraphAU} & IJCAI 22 & \underline{65.5} & 62.4 \\ 

BG-AU~\cite{BG-AU} & CVPR 23 & 64.1 & 58.2 \\

KS~\cite{KS} & ICCV 23 & 64.7 & 62.8 \\

FG-Net~\cite{FG-Net} & WACV 24 & 64.3 & \underline{65.4} \\

\midrule

\rowcolor{gray!30}
\textbf{AUFormer (Ours)} & - & \textbf{66.2} & \textbf{66.4} \\

\bottomrule[1pt]
\end{tabular}
\vspace{-2.5em}
\end{wraptable}

\textbf{Within-domain Evaluation.}
We first conduct within-domain evaluations on the widely used BP4D and DISFA, and compare AUFormer with thirteen state-of-the-art methods. Table \ref{tab:BP4D&DISFA} presents the average F1-score results of different methods and it can be observed that AUFormer achieves at least 0.7\% and 1.0\% improvement on BP4D and DISFA, respectively. The results for each individual AU are detailed in Appendix. Even though some methods like J{\^A}A-Net, KDSRL, FG-Net, BG-AU utilized additional auxiliary information such as landmark coordinates, large-scale relevant facial data, or biomechanical information, AUFormer demonstrates excellent performance without requiring extra relevant data. Besides, unlike most previous methods (such as J{\^A}A-Net, LP-Net, FAUDT, PIAP, etc.) that initialized models with well-trained parameters from BP4D when training on the DISFA, we train AUFormer from scratch solely on DISFA, achieving superior results. These advancements can be attributed to the proposed parameter-efficient collaboration mechanism, the tailored MoKE considering AU characteristics, and the meticulously crafted MDWA-Loss, which distinguishes AU difficulty and enables the discarding of potentially mislabeled samples. 

\textbf{Cross-domain Evaluation.}
To explore the generalization of AUFormer, we conduct bidirectional cross-domain evaluations, specifically from BP4D to DISFA and from DISFA to BP4D. Similar to \cite{FG-Net}, we utilize two folds of data from the source domain as the training set, while the entire dataset from the target domain is used as the testing set. We compare AUFormer with seven state-of-the-art methods on five AUs that are present in both datasets and the F1-score results of cross-domain evaluations are shown in Table \ref{tab:CrossDomain}. It can be observed that AUFormer outperforms all methods by at least 7.1\% in the DISFA to BP4D direction and is second only to FG-Net in the BP4D to DISFA direction, demonstrating its highly competitive generalization. For the BP4D to DISFA direction, we speculate that the higher occurrence rates of individual AUs in the BP4D (at least 15\%) lead AUFormer to capture more dataset-specific features, making it difficult to transfer seamlessly to DISFA. In contrast, FG-Net utilized features from the StyleGAN2~\cite{StyleGANDecoder,StyleGANEncoder} pre-trained on FFHQ~\cite{FFHQ}, a dataset with high-quality diverse facial data on a large scale, reducing the degree of overfitting on BP4D. Nevertheless, even without using more diverse additional data, AUFormer still outperforms other methods by at least 3.9\% in the BP4D to DISFA direction, except for FG-Net.

\begin{table}[t]
\caption{F1-scores (in \%) for cross-domain evaluations between BP4D and DISFA. The best and second best results for each column are bolded and underlined, respectively. $^*$The numbers are derived from a replication of \cite{FG-Net} based on open-source code.
}
\vspace{-0.8em}
\scriptsize
\centering

\begin{tabular}{l|ccccc|c|ccccc|c}
\toprule

Direction & \multicolumn{6}{c|}{BP4D $\rightarrow$ DISFA} & \multicolumn{6}{c}{DISFA $\rightarrow$ BP4D} \\
\midrule

AU & 1 & 2 & 4 & 6 & 12 & \textbf{Avg.} & 1 & 2 & 4 & 6 & 12 & \textbf{Avg.} \\
\midrule
DRML$^*$ \cite{DRML} & 10.4 & 7.0 & 16.9 & 14.4 & 22.0 & 14.1 & 19.4 & 16.9 & 22.4 & \underline{58.0} & 64.5 & 36.3  \\
J{\^A}A-Net$^*$ \cite{JAA-Net} & 12.5 & 13.2 & 27.6 & 19.2 & 46.7 & 23.8 & 10.9 & 6.7 & \underline{42.4} & 52.9 & \underline{68.3} & 36.2  \\
ME-GraphAU$^*$ \cite{ME-GraphAU} & 43.3 & 22.5 & 41.7 & 23.0 & 34.9 & 33.1 & 36.5 & 30.3 & 35.8 & 48.8 & 62.2 & 42.7  \\

Patch-MCD \cite{Patch-MCD} & 34.3 & 16.6 & \textbf{52.1} & 33.5 & 50.4 & 37.4 & - & - & - & - & - & -  \\
BG-AU~\cite{BG-AU} & - & - & - & - & - & 38.8 & - & - & - & - & - & -  \\
IdenNet \cite{IdenNet} & 20.1 & 25.5 & 37.3 & \textbf{49.6} & \textbf{66.1} & 39.7 & - & - & - & - & - & -  \\
FG-Net~\cite{FG-Net}  & \textbf{61.3} & \textbf{70.5} & 36.3 & \underline{42.2} & \underline{61.5} & \textbf{54.4} & \textbf{51.4} & \textbf{46.0} & 36.0 & 49.6 & 61.8 & \underline{49.0} \\
\midrule
\rowcolor{Gray!30}
\textbf{AUFormer (Ours)} & \underline{49.7} & \underline{49.9} & \underline{43.2} & 31.5 & 43.5 & \underline{43.6} & \underline{49.8} & \underline{43.8} & \textbf{45.7} & \textbf{66.6} & \textbf{74.5} & \textbf{56.1}  \\

\bottomrule
\end{tabular}
\label{tab:CrossDomain}
\vspace{-2.5em}
\end{table}

\begin{wrapfigure}{r}{0.5\textwidth}
    \vspace{-2.1em}
    \centering
    \includegraphics[width=1.0\linewidth]{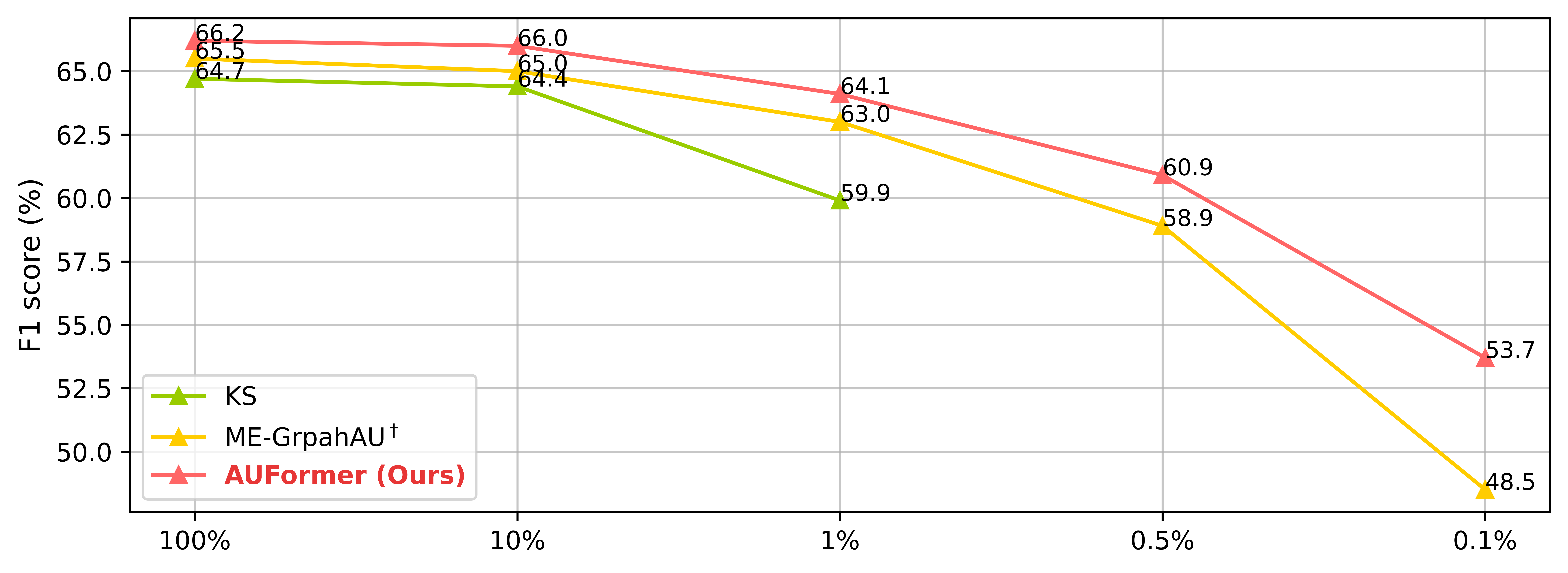}
    \vspace{-2.0em}
    \caption{
    Comparison of data efficiency capabilities between AUFormer, ME-GraphAU, and KS. $^\dagger$These numbers are derived from our replication based on open-source code.
    }
    \label{fig:DataEfficiency}
    \vspace{-2.5em}
\end{wrapfigure}

\setcounter{footnote}{0}

\textbf{Data Efficiency.}
We also investigate AUFormer's capability to efficiently utilize limited data on BP4D. Specifically, we perform equidistant sampling on the original training set to obtain a subset as the new training set, while keeping the original testing set unchanged for evaluation. We set the sampling rates to 10\%, 1\%, 0.5\%, and 0.1\%, respectively. Figure \ref{fig:DataEfficiency} illustrates the comparison of average F1-scores results of AUFormer with ME-GraphAU~\cite{ME-GraphAU} and KS~\cite{KS}, where the results of ME-GraphAU are obtained by reproducing the open-source code.\footnote{\url{https://github.com/CVI-SZU/ME-GraphAU}} It can be observed that when the sampling rate is 10\%, the performance remains relatively stable, which is due to the dense frame-level annotation of BP4D. As the sampling rate is further reduced, the performance decreases accordingly, which is an expected outcome. However, it should be noted that both ME-GraphAU and KS experience a drastic decline when the number of samples is extremely limited. In contrast, AUFormer shows a more gradual decline. Even with only 0.1\% of the original training set (less than 100 samples) utilized as training data, AUFormer achieves remarkably impressive results, showcasing its robust ability to extract effective information from limited data.

\textbf{Micro-expression Domain Evaluation.}
In addition to commonly used macro-expression AU datasets, we also consider the micro-expression AU dataset, namely CASME \Rmnum{2}~\cite{CASME}, to evaluate the ability of AUFormer in the micro-expression domain, which has been rarely explored in previous methods. We compare AUFormer with seven methods specifically designed for micro-expression AU detection, as well as ME-GraphAU, which performed excellently in macro-expression AU detection. 
\begin{table}[t]
\centering
\scriptsize
\caption{F1-score results (in \%) for micro-expression domain evaluation on CASME \Rmnum{2}. The best and second best results for each column are bolded and underlined, respectively. $^\dagger$These numbers are derived from our replication based on open-source code.}
\vspace{-1.2em}
\label{tab:CASME2}

\begin{tabular}{l|cccccccc|c}

\toprule[1pt]

\multirow{2}{*}{Method} & \multicolumn{8}{c|}{AU} & \multirow{2}{*}{\textbf{Avg.}} \\
\cmidrule(lr){2-9}
~ & 1 & 2 & 4 & 7 & 12 & 14 & 15 & 17 & ~ \\
\midrule

LBP-TOP~\cite{LBP-TOP} & 10.6 & 49.9 & 73.2 & 6.4 & 23.9 & 21.9 & 0.0 & 16.7 & 25.3 \\ 

LBP-SIP~\cite{LBP-SIP} & 23.1 & 38.9 & 73.5 & 8.9 & 21.4 & 29.8 & 43.2 & 42.9 & 35.2 \\

3DHOG~\cite{3DHOG} & 27.7 & 27.7 & 70.1 & 0.0 & 5.3 & 0.0 & 0.0 & 12.1 & 17.9 \\

SCA~\cite{SCA} & 28.6 & 45.3 & 88.8 & 24.7 & 47.9 & 33.3 & 39.5 & 51.6 & 45.0 \\

LKFS~\cite{LKFS} & 46.8 & 47.9 & 57.1 & \underline{51.6} & 55.3 & 50.7 & 47.5 & 47.8 & 50.6 \\

IICL~\cite{IICL} & 78.0 & 67.0 & \underline{89.0} & 30.0 & \underline{56.0} & 53.0 & 33.0 & 33.0 & 55.0 \\

ME-GraphAU$^\dagger$~\cite{ME-GraphAU} & \underline{79.3} & \textbf{85.8} & 88.3 & 35.3 & 39.4 & 47.9 & 49.5 & \underline{73.9} & 62.4 \\

ASP~\cite{ASP} & 76.6 & 71.7 & 81.8 & \textbf{53.5} & 55.3 & \textbf{68.1} & \underline{60.9} & 63.6 & \underline{66.4} \\

\midrule

\rowcolor{gray!30}
\textbf{AUFormer (Ours)} & \textbf{80.5} & \underline{81.9} & \textbf{92.2} & 39.2 & \textbf{66.0} & \underline{62.7} & \textbf{63.1} & \textbf{77.3} & \textbf{70.4} \\

\bottomrule[1pt]
\end{tabular}
\vspace{-2.2em}
\end{table} 
Table \ref{tab:CASME2} shows F1-score results of these methods on CASME \Rmnum{2}, where the results of ME-GraphAU are obtained by reproducing the open-source code. It can be seen that, although ME-GraphAU outperforms most methods specifically designed for micro-expression AU detection, demonstrating its feasibility in the micro-expression domain, it still falls short of ASP. On the other hand, AUFormer surpasses all methods with at least a 4\% improvement, proving that it can excel in the micro-expression domain as well. We hope that AUFormer can serve as a catalyst, inspiring more research that bridges the macro-expression and micro-expression domains.

\vspace{-0.9em}
\subsection{Ablation Study}
\label{sec:AblationStudy}

\begin{wraptable}{r}{0.55\textwidth}
    \tiny
    \centering
    \vspace{-5.1em}
    \caption{The average F1-score (in \%) results of various combinations of AUFormer components on BP4D. The collaboration mechanism is abbreviated as Collab.}
    \label{tab: AblationStudy}
    \vspace{-0.15em}
    
    \begin{tabular}{ccccccc|c}
    
    \toprule[1pt]
    
    \multirow{2}{*}{Baseline} & \multirow{2}{*}{PETL} & \multirow{2}{*}{Collab.} & \multicolumn{2}{c}{MoKE} & \multicolumn{2}{c|}{MDWA} & \multirow{2}{*}{\textbf{Avg.}} \\
    \cmidrule(lr){4-5}
    \cmidrule(lr){6-7}
     ~ & ~ & ~ & \multicolumn{1}{p{0.9cm}}{\centering MRF} & \multicolumn{1}{p{0.6cm}}{\centering CA} & \multicolumn{1}{p{0.5cm}}{\centering $\gamma$} & \multicolumn{1}{p{0.5cm}|}{\centering $m$} & ~ \\
    \midrule
    \cmark & ~ & ~ & ~ & ~ & ~ & ~ & 61.9 \\
    \cmark & ~ & ~ & ~ & ~ & \cmark & \cmark & 63.4 \\
    \cmark & \cmark & ~ & ~ & ~ & ~ & ~ & 63.6 \\
    \cmark & \cmark & \cmark & ~ & ~ & ~ & ~ & 64.3 \\
    \cmark & \cmark & \cmark & ~ & ~ & \cmark & \cmark & 65.2 \\
    \cmark & \cmark & ~ & \cmark & \cmark & ~ & ~ & 64.5 \\
    \cmark & \cmark & \cmark & \cmark & ~ & ~ & ~ & 64.8 \\
    \cmark & \cmark & \cmark & ~ & \cmark & ~ & ~ & 64.7 \\
    \cmark & \cmark & \cmark & \cmark & \cmark & ~ & ~ & 65.1 \\
    \cmark & \cmark & \cmark & \cmark & \cmark & \cmark & ~ & 65.6 \\
    \cmark & \cmark & \cmark & \cmark & \cmark & \cmark & \cmark & 66.2 \\
    
    \bottomrule[1pt]
    \end{tabular}
\vspace{-2.5em}
\end{wraptable}

We conduct ablation studies on BP4D dataset using subject-exclusive 3-fold cross-validation to explore the effects of each component of AUFormer, as well as the selection of important parameters and settings. Table \ref{tab: AblationStudy} presents the average F1-score results of \textit{various combinations of AUFormer's components}. We regard ViT/B-16 fully fine-tuned under the supervision of WA-Loss and WDI-Loss as the baseline, and its result is shown in the first row of Table \ref{tab: AblationStudy}.
\textbf{PETL paradigm.} 
Compared to fully fine-tuning, PETL paradigm offers a more promising approach to efficiently utilize scarce AU-annotated data and the third row of Table \ref{tab: AblationStudy} illustrates that it can lead to a 1.7\% improvement. Here, we default to choosing Convpass as the adaptation module for PETL, which is based on our exploration, with details shown in Table ~\ref{tab:TypeOfAdaptationModule}.
\textbf{Collaboration mechanism.} 
To better focus on the personalized characteristics of each AU, AUFormer develops a collaboration mechanism, which assigns dedicated experts to each AU. An expert for a specific AU collaborates with other AU experts from the current generation to integrate knowledge and passes down personalized knowledge to the next generation of expert corresponding to the same AU. The fourth row of Table \ref{tab: AblationStudy} highlights the superiority of collaboration mechanism over learning messy features of various AUs using a single Convpass.
\textbf{MoKE.} MoKE is specifically designed to capture knowledge crucial for AU detection, primarily encompassing the MRF and CA operators. The effectiveness of the MRF and CA operators when acting independently is demonstrated in the seventh and eighth rows of Table \ref{tab: AblationStudy}, respectively, verifying the necessity of multi-scale and correlation knowledge. Moreover, the sixth and ninth rows of Table \ref{tab: AblationStudy} also indicate that the joint action of the two operators leads to further improvement.
\textbf{MDWA-Loss.}
Compared to WA-Loss, MDWA-Loss can pay more attention to activated AU, differentiate the difficulty among unactivated AUs, and discard potentially mislabeled samples. We first replace the supervision signal from WA-Loss with MDWA-Loss, demonstrating the superiority of MDWA-Loss, as shown in the second and fifth rows of Table \ref{tab: AblationStudy}. Then, we investigate the roles of $\gamma$ and $m$ separately, showing gains of 0.5\% and 0.6\%, as demonstrated in the last two rows of Table \ref{tab: AblationStudy}.
\textbf{Mutual enhancement between MoKE and MDWA-Loss.}
Furthermore, it can be observed from the fourth, fifth, eighth, and eleventh rows of Table \ref{tab: AblationStudy} that when MoKE and MDWA-Loss are applied individually, they bring about gains of 0.8\% and 0.9\%, respectively, while when they act together, they lead to improvements of 1.0\% and 1.1\%, showcasing the synergistic effect between them. The underlying reason is that MDWA-Loss can guide MoKE to learn more precise features, and in turn, MoKE stimulates MDWA-Loss to propagate more effective gradients.

\begin{table}[t]
\tiny
\centering
\caption{The average F1-score (in \%) on BP4D with different parameter or setting choices. Default settings are marked in \colorbox{baselinecolor}{gray}.}
\vspace{-1.0em}
\label{tab:hyperparameter_1}
\subfloat[
Type of adaptation modules.
\label{tab:TypeOfAdaptationModule}
]{
\centering
\begin{minipage}{0.21\linewidth}{\begin{center}
\begin{tabular}{x{40}|x{25}}
\toprule[1pt]
Type        &   Avg.            \\
\midrule
Adapter     &   63.1            \\
LoRA        &   63.0            \\
VPT         &   62.6            \\
\rowcolor{gray!30}
Convpass    &   \textbf{63.6}   \\
\bottomrule[1pt]
\end{tabular}
\end{center}}\end{minipage}
}
\hspace{0.3em}
\subfloat[
Dilated rates in MRF.
\label{tab:DilationRateInMRF}
]{
\begin{minipage}{0.25\linewidth}{\begin{center}
\begin{tabular}{x{45}|x{25}}
\toprule[1pt]
Dilated rates   &   Avg.            \\
\midrule
\rowcolor{gray!30}
1, 3, 5         &   \textbf{66.2}   \\
1, 3            &   65.9            \\
1               &   65.6            \\
1, 2, 3         &   65.9            \\
\bottomrule[1pt]
\end{tabular}
\end{center}}\end{minipage}
}
\hspace{0.3em}
\subfloat[
Size of correlation neighborhood in CA.
\label{tab:CorrelationNeighborhoodInCA}
]{
\begin{minipage}{0.21\linewidth}{\begin{center}
\begin{tabular}{x{40}|x{25}}
\toprule[1pt]
$S \times S$    &   Avg.            \\
\midrule
\rowcolor{gray!30}
$3 \times 3$    &   \textbf{66.2}   \\
$5 \times 5$    &   65.9            \\
$7 \times 7$    &   66.0            \\
\bottomrule[1pt]
\multicolumn{2}{c}{~}               \\
\end{tabular}
\end{center}}\end{minipage}
}
\hspace{0.3em}
\subfloat[
Truncation margin $m$ in MDWA-Loss.
\label{tab:TruncatedMargin}
]{
\begin{minipage}{0.21\linewidth}{\begin{center}
\begin{tabular}{x{40}|x{25}}
\toprule[1pt]
$m$     &   Avg.            \\
\midrule
0.01    &   65.6            \\
0.05    &   65.9            \\
\rowcolor{gray!30}
0.1     &   \textbf{66.2}   \\
0.2     &   65.4            \\
\bottomrule[1pt]
\end{tabular}
\end{center}}\end{minipage}
}
\vspace{-5.5em}
\end{table}


Table \ref{tab:hyperparameter_1} presents the average F1-score results for \textit{different choices of important parameters or settings.}  
\textbf{Adaptation modules.} We explore the effects of several representative adaptation modules in PETL paradigm in AU detection, including Adapter~\cite{Adapter}, LoRA~\cite{LoRA}, VPT~\cite{VPT}, and Convpass~\cite{Convpass}, as shown in Table \ref{tab:TypeOfAdaptationModule}. It can be seen that VPT performs the worst, Adapter and LoRA are in the middle, and Convpass is the optimal. This is because the convolutional layers introduced in Convpass align with the spatial information requirements in AU detection. Therefore, when designing MoKE, we also incorporate convolutional layers. 
\textbf{Dilated rates.} 
We default the dilated rates $r_1$, $r_2$ and $r_3$ to 1, 3, and 5, and explore the impact of different dilated rates on performance, as shown in Table \ref{tab:DilationRateInMRF}. When removing the convolutional layers with dilated rates of 5 or 3 and 5, the performance experiences varying degrees of decline due to the reduced receptive field. Interestingly, experimenting with dilated rates of 1, 2, and 3 shows equivalent performance to rates 1 and 3, indicating the convolutional layers with a dilated rate of 2 does not bring an information increment, affirming the necessity of introducing convolutional layers with larger dilated rates.
\textbf{Correlation neighborhood size.} 
How much contextual information to aggregate within the correlation neighborhood is a question to consider. We set the neighborhood size $S \times S$ to $3 \times 3$, $5 \times 5$, and $7 \times 7$, and observe that a larger neighborhood did not lead to a better result, as seen in Table \ref{tab:CorrelationNeighborhoodInCA}. Therefore, we opt for the more computationally efficient $3 \times 3$.
\textbf{Truncation margin.}
We investigate the setting of the truncation margin $m$, as shown in Table \ref{tab:TruncatedMargin}. When $m$ is too small, MDWA-Loss almost fails to serve its purpose of discarding potentially mislabeled samples. Conversely, when $m$ is too large, MDWA-Loss may lead to misjudgment, ignoring samples that are actually correctly labeled. In summary, we set $m$ to 0.1. In addition to the above mentioned, we also explore the choices of adaptation module placement, reduced channel number $d$ in MoKE, correlation matrix calculation method, boundaries of $\gamma$, and applicability of AUFormer on various backbones~\cite{Swin}, etc. For more details, please refer to Appendix.

\begin{figure}[t]
  \centering
  \begin{minipage}[b]{0.43\linewidth}
      \begin{minipage}{0.45\textwidth}
        \centering
        \includegraphics[width=0.88\linewidth]{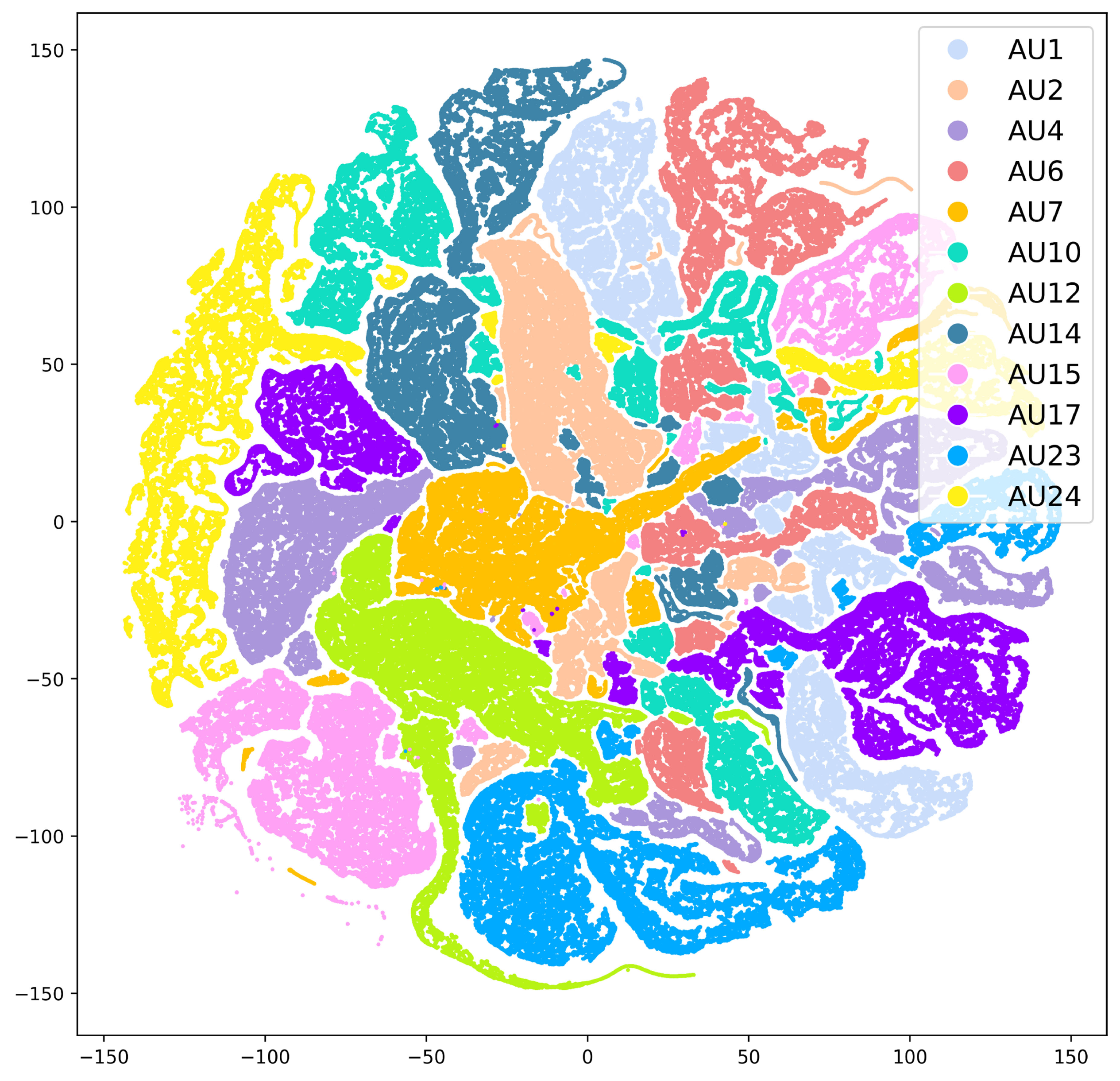}
        \subcaption{w/o $\mathcal{L}_\text{MDWA}^{\prime}$}
        \label{fig:withoutLMoKE}
      \end{minipage}
      \begin{minipage}{0.45\textwidth}
        \centering
        \includegraphics[width=0.88\linewidth]{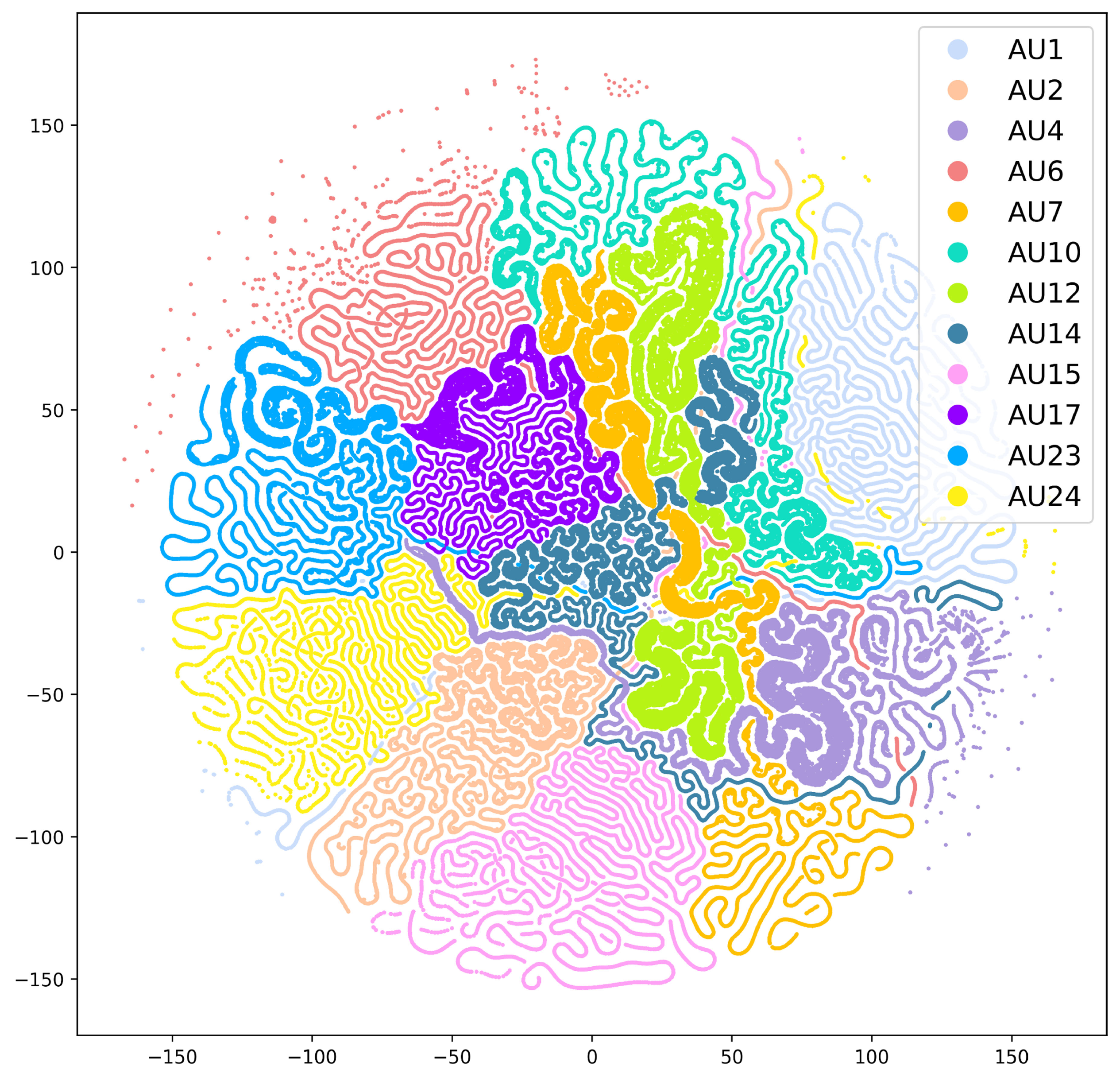}
        \subcaption{w/ $\mathcal{L}_\text{MDWA}^{\prime}$}
        \label{fig:withLMoKE}
      \end{minipage}
      \vspace{-0.5em}
      \caption{The t-SNE~\cite{t-SNE} distribution of the [CLS] tokens from the final group of MoKEs.}
    \label{fig:tsne}
  \end{minipage}
  \hspace{0.3em}
  \begin{minipage}[b]{0.53\linewidth}
    \centering
    \includegraphics[width=0.85\linewidth]{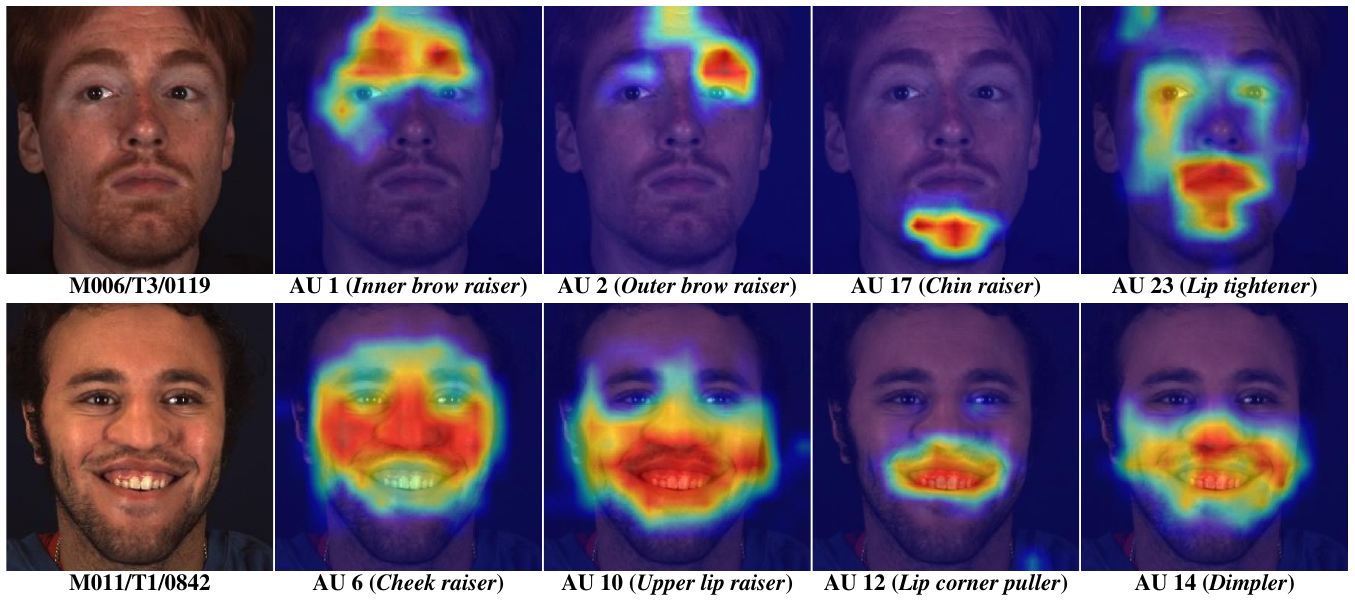}
    \vspace{-0.5em}
    \caption{Activation maps for MoKEs corresponding to activated AUs generated using Grad-CAM~\cite{Grad-CAM}.}
    \label{CAM}
  \end{minipage}
  \vspace{-2.3em}
\end{figure}

\textbf{Visualization.}
The question of whether MoKEs targeting different AUs actually capture the personalized features of corresponding AUs is a matter of concern, and we illustrate this through visualization on BP4D. We first visualize the distribution of [CLS] tokens from the final group of MoKEs using t-SNE to explore the necessity of setting $\mathcal{L}_\text{MDWA}^{\prime}$. As shown in Figure \ref{fig:withoutLMoKE}, without supervision of $\mathcal{L}_\text{MDWA}^{\prime}$ for MoKEs, their outputs are chaotic and disorderly. In contrast, when $\mathcal{L}_\text{MDWA}^{\prime}$ is introduced, the outputs from the same MoKE corresponding to a certain AU are clustered together, distinct from the outputs of other MoKEs, as shown in Figure \ref{fig:withLMoKE}. Then, we generate activation maps for activated AUs utilizing Grad-CAM to observe the facial regions focused on by individual MoKEs. Figure \ref{CAM} illustrates that each MoKE indeed emphasizes regions related to their respective AU. For example, MoKEs corresponding to AU1 (Inner brow raiser) and AU6 (Cheek raiser) place their focal points on the inner brow and cheeks, respectively. More activation maps are detailed in Appendix.

\section{Conclusion and Future Work}

\label{sec:Conclusion}
In this paper, we investigate the application of PETL paradigm in AU detection for the first time. We propose AUFormer and develop a novel MoKE collaboration mechanism for efficiently leveraging the pre-trained ViT. We also introduce a novel MDWA-Loss, which considers the properties of AU datasets and difficulty differences in recognizing different AUs. We demonstrate the state-of-the-art performance of AUFormer from multiple perspectives. 
We note that the study on employing ViT in AU detection is still at an early stage. Future directions include: (\rmnum{1}) exploring a dynamic collaborative mechanism that can adaptively adjust the contributions of each MoKE. (\rmnum{2}) utilizing the personalized features of each AU extracted by MoKEs for further subsequent processing.


%
%
\bibliographystyle{splncs04}
\bibliography{main}
\end{document}